\documentclass[sigconf]{acmart}

\copyrightyear{2023}
\acmYear{2023}
\setcopyright{acmlicensed}
\acmConference[MM '23]{Proceedings of the 31st ACM International Conference on Multimedia}{October 29-November 3, 2023}{Ottawa, ON, Canada}
\acmBooktitle{Proceedings of the 31st ACM International Conference on Multimedia (MM '23), October 29-November 3, 2023, Ottawa, ON, Canada}
\acmPrice{15.00}
\acmDOI{10.1145/3581783.3611732}
\acmISBN{979-8-4007-0108-5/23/10}

\usepackage{xcolor}
\colorlet{dark-blue}{blue!50!black}
\colorlet{dark-cyan}{cyan!75!black}
\colorlet{dark-purple}{purple!50!black}
\colorlet{dark-red}{red!75!black}
\colorlet{dark-green}{green!75!black}
\colorlet{dark-orange}{orange!50!black}
\colorlet{dark-gray}{black!75}
\colorlet{light-gray}{black!30}
\definecolor{nice-red}{HTML}{E41A1C}
\definecolor{nice-orange}{HTML}{FF7F00}
\definecolor{nice-yellow}{HTML}{FFC020}
\definecolor{nice-green}{HTML}{39b54a}
\definecolor{nice-blue}{HTML}{0071bc}
\definecolor{nice-purple}{HTML}{984EA3}

\definecolor{darkGreen}{rgb}{0, 0.6, 0}
\definecolor{darkRed}{rgb}{0.9, 0, 0} 
\definecolor{cyan}{rgb}{0, 0.5, 0.6} 
\definecolor{darkViolet}{rgb}{0.58, 0, 0.83}

\definecolor{lightgreen}{rgb}{0.4, .9, 0.4}
\definecolor{lightred}{rgb}{1, 0.5, 0.51}
\definecolor{xgray}{rgb}{0.6, 0.6, 0.6}
\definecolor{Highlight}{HTML}{39b54a}
\definecolor{citecolor}{HTML}{0071bc}

\usepackage{graphicx}
\usepackage{subcaption}
\usepackage{float}
\usepackage{caption}
\usepackage{lscape}                                         
\usepackage{wrapfig}
\usepackage{balance}
\usepackage{chngcntr}

\usepackage[lined,ruled,linesnumbered]{algorithm2e}
\usepackage{animate} 

\usepackage{booktabs}                   
\usepackage{multirow}

\usepackage{bm}                          

\usepackage{amsmath}  

\usepackage{units}
\usepackage{color}
\usepackage[T1]{fontenc}    
\usepackage{amsfonts}       
\usepackage[utf8]{inputenc} 
\usepackage{nicefrac}       
\usepackage{microtype}      
\usepackage{pifont}         
\usepackage{comment}

\usepackage[mathscr]{euscript}
\usepackage{colortbl}
\usepackage{diagbox}

\newcommand{\tablestyle}[2]{\setlength{\tabcolsep}{#1}\renewcommand{\arraystretch}{#2}\centering\small}

\newcommand{\mycaption}[2]{\caption{\textbf{#1.}\xspace#2}}

\def\ie{\textit{i.e.,\ }}
\def\eg{\textit{e.g.,\ }}

\newcommand{\heading}[1]{\noindent\textbf{#1}}

\newcolumntype{x}[1]{>{\centering\arraybackslash}p{#1pt}}
\newcolumntype{y}[1]{>{\raggedright\arraybackslash}p{#1pt}}
\newcolumntype{z}[1]{>{\raggedleft\arraybackslash}p{#1pt}}

\newlength\savewidth\newcommand\shline{\noalign{\global\savewidth\arrayrulewidth
  \global\arrayrulewidth .8pt}\hline\noalign{\global\arrayrulewidth\savewidth}}

\newcommand{\figref}[1]{Figure~\ref{fig:#1}}
\newcommand{\tabref}[1]{Table~\ref{tab:#1}} 

\newcommand{\ignore}[1]{}   

\newcommand{\cmark}{{\ding{51}}}
\newcommand{\xmark}{{\ding{55}}}

\graphicspath{{figure/}}
\graphicspath{{image/}}

\AtBeginDocument{%
  \providecommand\BibTeX{{%
    \normalfont B\kern-0.5em{\scshape i\kern-0.25em b}\kern-0.8em\TeX}}}

\acmSubmissionID{187}

    \makeatletter
\def\@fnsymbol#1{\ensuremath{\ifcase#1\or \color{red}{\dagger}\or \ddagger\or
   \mathsection\or \mathparagraph\or \|\or **\or \dagger\dagger
   \or \ddagger\ddagger \else\@ctrerr\fi}}
    \makeatother

\begin{document}

\title{Beyond Domain Gap: Exploiting Subjectivity in Sketch-Based Person Retrieval}

\author{Kejun Lin}
\email{2019302110249@whu.edu.cn}
\orcid{0009-0001-8900-8396}
\affiliation{%
  \institution{Wuhan University}
  \city{Wuhan}
  \country{China}
}

\author{Zhixiang Wang}
\authornotemark[1]
\orcid{0000-0002-5016-587X}
\email{wangzhixiang@g.ecc.u-tokyo.ac.jp}
\affiliation{%
  \institution{The University of Tokyo}
  \city{Tokyo}
  \country{Japan}
}

\author{Zheng Wang}
\authornote{Corresponding authors: Zheng Wang, Zhixiang Wang}
\email{wangzwhu@whu.edu.cn}
\orcid{0000-0003-3846-9157}
\affiliation{%
  \institution{National Engineering Research Center for Multimedia Software, School of Computer Science, Wuhan University}
  \city{Wuhan}
  \country{China}
}

\author{Yinqiang Zheng}
\email{yqzheng@ai.u-tokyo.ac.jp}
\orcid{0000-0001-7434-5069}
\affiliation{%
  \institution{The University of Tokyo}
  \city{Tokyo}
  \country{Japan}
}

\author{Shin'ichi Satoh}
\email{satoh@nii.ac.jp}
\orcid{0000-0001-6995-6447}
\affiliation{%
  \institution{National Institute of Informatics}
  \city{Tokyo}
  \country{Japan}
}

\begin{abstract}
Person re-identification (re-ID) requires densely distributed cameras. In practice, the person of interest may not be captured by cameras and therefore need to be retrieved using subjective information (e.g., sketches from witnesses). Previous research defines this case using the sketch as sketch re-identification (Sketch re-ID) and focuses on eliminating the domain gap. Actually, \textbf{subjectivity} is another significant challenge. 
We model and investigate it by posing a new dataset with multi-witness descriptions. It features two aspects. 1)~Large-scale. It contains over 4,763 sketches and 32,668 photos, making it the largest Sketch re-ID dataset. 2)~Multi-perspective and multi-style. 
Our dataset offers multiple sketches for each identity. Witnesses' subjective cognition provides multiple perspectives on the same individual, while different artists' drawing styles provide variation in sketch styles.
We further have two novel designs to alleviate the challenge of subjectivity. 1) Fusing subjectivity. We propose a non-local (NL) fusion module that gathers sketches from different witnesses for the same identity. 2) Introducing objectivity. An AttrAlign module utilizes attributes as an implicit mask to align cross-domain features. To push forward the advance of Sketch re-ID, we set three benchmarks (large-scale, multi-style, cross-style). Extensive experiments demonstrate our leading performance in these benchmarks. Dataset and Codes are publicly available at: \url{https://github.com/Lin-Kayla/subjectivity-sketch-reid}
\end{abstract}




\maketitle

\section{Introduction}
\label{sec:introduction}

\begin{figure}[t]
    \centering
    \includegraphics[width=0.72\linewidth]{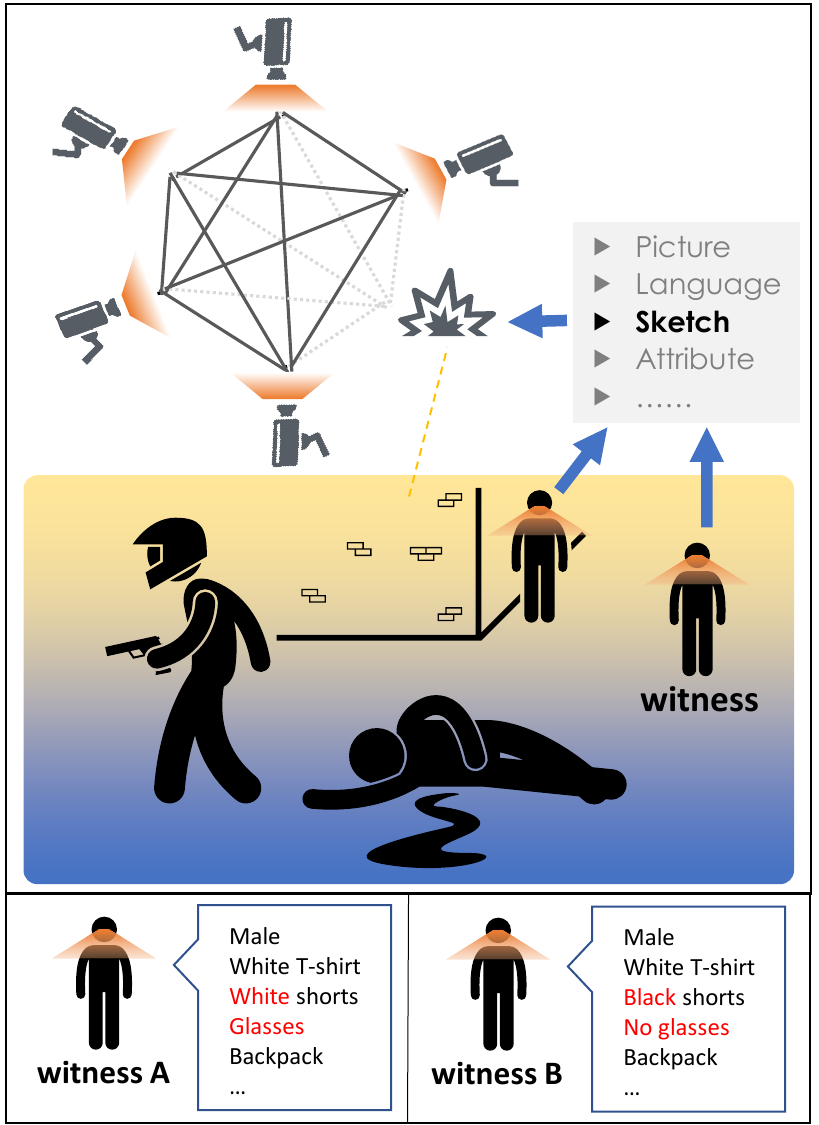}
    \mycaption{The problem}{The success of person re-identification system relies on densely distributed cameras. But, the system cannot always guarantee us a photo when an event happened. To make person retrieval feasible, we need to use subjective cues from witnesses as queries. Note that as witnesses have different observations/perceptions, that information may be partial and uncertain.}
    \label{fig:intro_problem} 
\end{figure}

Person re-identification (re-ID) is a promising technique for a lot of practical applications.
It aims to match a photo query of a person within a camera system database. Despite the great progress of re-ID, most research neglects 
the fact that the sparsely distributed camera system cannot always give a photo query when an event happened. 
To make person retrieval feasible again, several works take advantage of witnesses' cues as queries (Figure~\ref{fig:intro_problem}), \eg sketches~\cite{pang2018cross} and natural language descriptions~\cite{li2017person}. In this paper, we study the case of cross-domain re-ID using the sketch of a person's body appearance, \ie sketch re-identification (Sketch re-ID). 
As opposed to the traditional forensic facial sketches matching~\cite{kiani2012face}, it leverages \emph{whole-body} sketches to match against a photo gallery database.

Sketch re-ID faces several challenges due to the domain gap between sketches and photos, such as view and pose discrepancies, communication barriers between witnesses and artists, etc. It is also the main focus of previous works, \eg PKU-Sketch~\cite{pang2018cross}, which we noticed proposed the only Sketch re-ID dataset to date. However, the limited scale of the dataset has made it difficult to develop data-driven algorithms and conduct fair evaluations. A closer examination of sketch re-ID reveals that subjectivity is often neglected. Witness subjectivity, in particular, plays a critical role in sketch-based identification. According to the Boston Marathon Bombing case~\cite{feris2014attribute}, witnesses have different views and provide only partial descriptions, making it challenging to capture the complete appearance of the suspect. Moreover, their perceptions of the same scene may differ, and information gets lost in memories over time, making sketch information partial and uncertain. Additionally, artists have various drawing styles that can impact the accuracy of the sketches. While PKU-Sketch has considered artists' styles, their one-sketch-per-identity setting is inadequate for style generalization and leaves little room for future research to explore the impact of styles.

To address the subjectivity in witnesses' perception and artists' drawing, we collected a more practical dataset, Market-Sketch-1K. 
It contains the following distinctive features: \textbf{1)} \textit{Large-scale.} The size of our dataset is much larger than the prior PKU-Sketch dataset (4.7K vs 200).
\textbf{2)} \textit{Multi-perspective and multi-style.} Our dataset includes multiple sketches for each identity, each reflecting a different witness's subjective cognition and therefore providing multiple perspectives on the same individual. Each sketch is drawn by different artists, providing a variation in sketch styles.
By diversifying and enriching the dataset, we are able to achieve two main goals: \textbf{}1) explore the impact of multiple perspectives and sketch styles on person re-identification, and \textbf{2)} incorporate subjective data to build a more generalizable and robust algorithm.


\begin{figure*}[t]
    \centering
    \includegraphics[width=\linewidth]{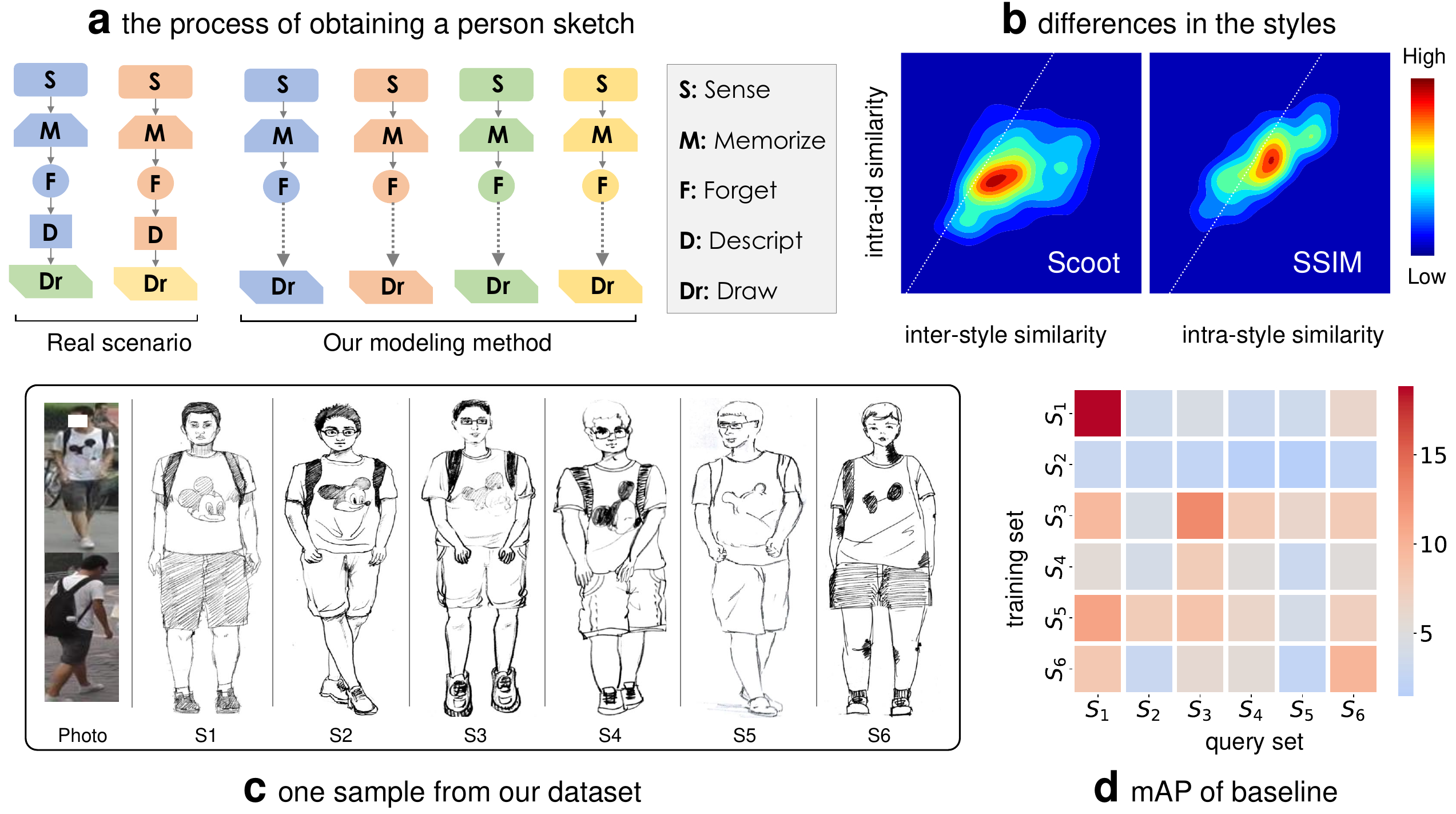}
\caption{
\textbf{Overview of our dataset.} 
\textbf{a}: the process of obtaining a sketch. 
After perceiving the target person from different viewpoints, the witness describes the details to an artist. As communication loss between witnesses and artists is not our main focus, in our process, the witness and the artist are the same people (denoted with the same color). We keep artists inaccessible to the photo during drawing (denoted by dashed lines) to guarantee their perspective descriptions, while PKU-Sketch does not enforce this. 
\textbf{b}: intra-style similarities are larger than intra-id similarities in terms of perceptual metrics SSIM~\cite{wang2004image} and Scoot~\cite{fan2019Scoot}. 
\textbf{c}: indicates a sample that we obtain inconsistent descriptions from different witnesses, and each artist has varied sketching styles. \textbf{d}: shows using sketches from the same artist to train a model and testing it using sketches with different artists is challenging.
    }
    \label{fig:intro_motivation}
\end{figure*}

We crack the difficulty caused by subjectivity with a scalable sketch re-ID system, which adapts to both single-query and multi-query scenarios and is flexible to the varying input sketch numbers. It comes with two effective designs: \textbf{1)} fusing multiple subjective queries from different witnesses with our proposed non-local(NL) fusion module. It filters out the noise in multiple sketches while keeping the long-distance dependencies available.
\textbf{2)} Introducing objective information as guidance (\ie accurate attributes) in the training phase. We present an AttrAlign module casting the attributes as an implicit mask to align subjective sketches and the corresponding photo. We find that the fusion module not only helps us deal with subjectivity but also facilitates style variation, allowing our model to generate well to an unseen sketch style or even a distinct dataset. 

With the proposed method and the dataset, we raise three benchmarks: \textbf{1)}~a large-scale benchmark to promote robust algorithms that address the domain discrepancies; \textbf{2)}~a new multi-style benchmark using multiple sketches of an identity as a query to study the possibility of boosting performances by employing available sketches. \textbf{3)}~a new cross-style benchmark verifying the generalization ability. Our method achieves favorable results on all of them.
The contributions of this paper are three aspects:

\begin{itemize}
    \item We deliver Market-Sketch-1K, a new Sketch re-ID dataset with characteristics that are large-scale, multi-perspective, and multi-style. It reveals that models trained with a sketch per identity are style-biased and not applicable to the case with different kinds of sketch styles.
    
    \item To alleviate the issue caused by subjectivity, we have two novel designs: non-local (NL) fusion and AttrAlign. NL fusion gathers sketches from different witnesses for the same identity. AttrAlign utilizes attributes as an implicit mask to align cross-domain features.

    \item With the proposed dataset, we set three benchmarks: a large-scale benchmark, a new multi-style benchmark, and a new cross-style benchmark. Our method achieves leading performance in these benchmarks.
    
\end{itemize}

\section{Related Work}
\label{sec:related}

\paragraph{Datasets for person re-ID} 
General re-ID aims to match photos captured in the daytime under different camera views. There are several public datasets, \eg VIPeR~\cite{gray2008viewpoint}, Market-1501~\cite{zheng2015scalable}, and MSMT17~\cite{wei2018person}. They are progressively constructed with a larger number of cameras and identities. 
Impressive results have been achieved on these datasets. 
There are also several works investigating the problem of the modality gap in re-ID. For example, \cite{wu2017rgb,dai2018cross,wang2019learning} attempted to address the modality gap between visible and infrared light photos, and their algorithms were evaluated on the SYSU-MM01~\cite{wu2017rgb} and RegDB~\cite{sensors17} datasets. \cite{li2017person,sarafianos2019adversarial} paid attention to the gap between natural language descriptions and photos, and algorithms were evaluated on the CUHK-PEDES~\cite{wu2017rgb} dataset. \cite{pang2018cross} is the first work to investigate the Sketch re-ID. 

\paragraph{Datasets for sketch-based image retrieval.} 
We categorize datasets into two categories. One is the \emph{category-level}, \eg Sketchy~\cite{sangkloy2016sketchy}, Sketch-250~\cite{eitz2012humans}, and PACS~\cite{li2017deeper}. 
This kind of dataset often ignores the style variations, since the influence of the style factor is relatively small and not fatal. The other category is the \emph{instance-level}, \eg QMUL-Shoe, QMUL-Chair~\cite{yu2016sketch}, and Handbag~\cite{song2017deep}. 
\cite{gao2012face,peng2018face} proposed face sketch datasets consisting of multiple stylistic sketches. However, they ignore the effect of the style factor. In particular, 
their attentions are the forensic facial sketch matching, and styles are almost consistent. 
Thus, most cross-domain or sketch-based datasets are concerned with the objective issue of cross-domain rather than the subjective issue of sketches being drawn by people.

\paragraph{Cross-modal Person Re-identification (Re-ID)} There're four main kinds of heterogeneous Re-ID~\cite{ye2021deep}, including Re-ID between depth and RGB images, text-to-image Re-ID, visible-infrared Re-ID and cross-resolution Re-ID. Sketch Re-ID resembles the most with visible-infrared Re-ID (VI-ReID). One common approach is using a two-stream network~\cite{ye2018hierarchical, ye2019bi} to model the modality-sharable and -specific information. Another branch is generative adversarial networks (GANs) - based methods~\cite{choi2020hi, dai2018cross, wang2019learning, wang2019rgb}. GANs capture discriminative factors for person Re-ID better, but they require a lot of parameters and heuristics to train the network.

Although sketch Re-ID shares similarity to VI-ReID, sketches differ significantly from infrared images in terms of style, level of detail, and representation. To address the distinct subjectivity issue in sketch Re-ID, we leverage objective information (attributes). We cast the attributes as an implicit mask to align the cross-domain features instead of generating the masks using the similarity between features~\cite{park2021learning}. Via the zero-shot transfer ability in CLIP~\cite{radford2021learning}, the masks carry more semantic meaning and, thus better alignment.

\paragraph{Zero-Shot Fine-Grained Sketch-Based Image-Retrieval (ZS-FG-SBIR)}
Fine-grained Sketch-Based Image-Retrieval(FG-SBIR) uses sketch as the query for \textit{instance-level} matching against a photo gallery. Early deep-learning based works include triplet-ranking based siamese network~\cite{yu2016sketch} and its augmented version, such as via attention with higher-order retrieval loss~\cite{song2017deep}, cross-domain image generation~\cite{pang2018cross} and text tags~\cite{song2017fine}. More recent works include hierarchical co-attention~\cite{sain2020cross}, reinforcement learning-based early retrieval~\cite{bhunia2020sketch}, etc.
These early works assume the dataset to be perfect, \ie a perfect depiction of the paired photo. A recent work~\cite{bhunia2022sketching} addresses the "sketchy" nature of sketches but focuses on amateur sketches' free-flow nature. Ours also recognizes the imperfection of a sketch, but we aim to address the subjectivity issue in sketches drawn by skilled artists.

Zero-shot Sketch-Based Image-Retrieval(ZS-SBIR) aims to generalize knowledge learned from seen training classes to unseen testing classes. Early works align sketch, photo and semantic features via adversarial training~\cite{dutta2019semantically}, gradient reversal layer~\cite{dey2019doodle}, etc. Further, they used graph convolution network~\cite{zhang2020zero}, knowledge distillation~\cite{tian2021relationship, wang2022prototype}, test-time training via reconstruction on test-set sketches~\cite{sain2022sketch3t}, etc.
Semantic transfer for ZS-SBIR has limited to using word embeddings only, either directly~\cite{dutta2019semantically, wang2021transferable, zhang2020zero} or indirectly~\cite{liu2019semantic, tian2021relationship}.

However as a non-trivial problem, ZS-FG-SBIR is less studied. A concurrent work~\cite{sain2023clip} leverages CLIP and prompt learning for semantic transfer via contrastive learning. Unlike them, we cast attributes as an implicit mask using CLIP for cross-domain alignment.

\section{Market-Sketch-1K}
\label{sec:data}

\paragraph{Modeling process.}

We build our Market-Sketch-1K based on the most popular re-ID dataset Market-1501~\cite{zheng2015scalable}, a large-scale dataset collected in an open system. Market-1501 contains 32,668 photos of 1,501 unique person identities. It acts as the photo part of our dataset. To build the sketch part, we invited 6 artists to draw each identity based on their cognition. The process of constructing the sketch part of our dataset is as follows: \textbf{1)} We randomly selected identities from the Market-1501 dataset as the reference subset. 
In general, we selected 498 identities from the training set of Market-1501, and 498 identities from the query set of Market-1501. 
\textbf{2)} For each artist, we showed him/her randomly selected photos of each identity, which vary in viewpoint, lighting, pose, etc. The variation compels the artists to form a general perception of the target person, instead of simply memorizing the picture.
After viewing the photos, the artist drew a sketch of the person without any communication. 
Note that artists drew sketches not by imitating the photos but by their subjective cognition and judgment. 
Since artists looked at different photos of each identity, the sketches would contain artists' subjective cognition and their painting styles. 
\textbf{3)} After collecting all sketches, we scanned them into the electronic version, verified manually, cropped the papers' white margins, down-sampled the sketch images to the same scale, enhanced the sketch lines, and finally archived created sketches according to their identities and artist IDs. Different from free-hand sketches, forensic sketches often use densely distributed lines and varying stroke widths to simulate the lighting, texture, etc. for more accurate representation. These characteristics make it difficult to produce vector-like data, so we follow the majority to only produce pixel-based data. 
In total, Market-Sketch-1K consists of 4,763 sketches of 996 identities and 32,668 photos of 1,501 identities. \figref{intro_motivation}{a} demonstrates the feature of our process of obtaining a sketch.

\paragraph{Two distinctive features.}
\emph{Large-scale.}  
More diverse and abundant sketches make our dataset the largest Sketch re-ID dataset to date. 
Compared with those for face sketch retrieval and Sketch re-ID, our dataset excels at the number of identities, styles, camera views, etc. as shown in 
the appendix. 
\emph{Multi-perspective and multi-style.}  
To investigate the characteristics of sketches, we used two perceptual metrics, SSIM~\cite{wang2004image} and Scoot~\cite{fan2019Scoot}, to calculate the similarity between every two sketches. We conducted the calculation of 100 trials and got the distributions of sketch similarities and the average values of similarities according to different artists. The figure shows that, in most cases, intra-style similarities of different sketches are larger than intra-id similarities (\figref{intro_motivation}{b}), indicating that it's easier to recognize the artist than the target person. We categorize this variation into three aspects: a) the discrepancy of artists' viewpoints due to the different photos shown to them; b) different cognitive abilities because of artists' different backgrounds; c) artists' different painting styles.


\paragraph{Impact of perspective and style variations.}
We carried out preliminary studies based on the cross-modal AGW baseline~\cite{ye2021deep}. The results are shown in \figref{intro_motivation}(d). 
The data from row $S_i$ and column $S_j$ indicates the mean Average Precision (mAP) of training on $S_i$ and testing on $s_j$. We can see that under most styles, training and testing on the same style perform better, although all perform under par.
The results show that performances are affected heavily by the sketch qualities (the results of $S_1$, $S_3$, $S_5$, and $S_6$, which have the same training and testing identities and setting).

\subsection{Benchmark design}
\paragraph{Dataset splits.} 
Sketches are divided into 6 different groups ``$S_1$--$S_6$'' according to sketch styles/artists. \tabref{exp_datasplit} shows our dataset splits. 
In the training stage, sketches of selected groups and all photos (the same as Market-1501) are used. The training set is denoted as $T = \{x_p^k,x_s^k,y^k|y^k \in \mathcal{Y}^{tr}\}$. In the test stage, sketches of selected groups are considered as the query $Q = \{x_s^k, y^k|y^k \in \mathcal{Y}^{qr}\}$ and photos act as the gallery $G = \{x_p^k, y^k|y^k \in \mathcal{Y}^{gr}\}$. $\mathcal{Y}^{tr},\mathcal{Y}^{qr},\mathcal{Y}^{gr}$ denote the training, query and gallery identities separately. We adopt the zero-shot setting~\cite{zheng2015scalable}, which means the training and testing sets share no identical target person (\ie $\mathcal{Y}^{tr}\!\cap\! \mathcal{Y}^{qr} \!=\! \emptyset$).



\paragraph{Benchmark.} There are three challenges in our dataset, including \textbf{1)} the appearance and domain discrepancies, \textbf{2)} the perspective and style variations, and \textbf{3)} multiple available sketches for each query. Thus, we set three benchmarks, \ie single-query, multi-query, and cross-style retrieval. In particular, we evaluate the algorithm using multiple sketches of each identity together as a query (multi-query). In practical situations, multiple artists may join the retrieval task together. They saw the target person from different views, and they have different ideas and painting styles. This inspires up to raise this new benchmark. 

%


\begin{table}[!t]
     \centering
    \caption{\textbf{Dataset splits.} During testing, we use the sketch as our query and the photo counterpart as our gallery.}
     \label{tab:exp_datasplit}
    \resizebox{1\linewidth}{!}{
    \tablestyle{6pt}{1.0}
    \begin{tabular}{l|rrrrrr|r}
    & \multicolumn{6}{c|}{\textbf{\#sketch}} & \textbf{\#photo}\\
    & $S_1$  & $S_2$ & $S_3$  & $S_4$  & $S_5$  & $S_6$ &  \\
    \shline
     Train  &  498 & 93 & 498 & 247 & 498 & 498 & 12,936\\
     Test   &   498 & 130 & 498 & 253 & 498 & 498 & 19,732\\
    \end{tabular}}
\end{table}

\paragraph{Evaluation metrics.} We employ the standard Cumulative Matching Characteristics (CMC) and mean Average Precision (mAP)~\cite{zheng2015scalable} as our metrics. 

\section{Method}


\begin{figure*}[!tb]
    \centering
    \includegraphics[width=\linewidth]{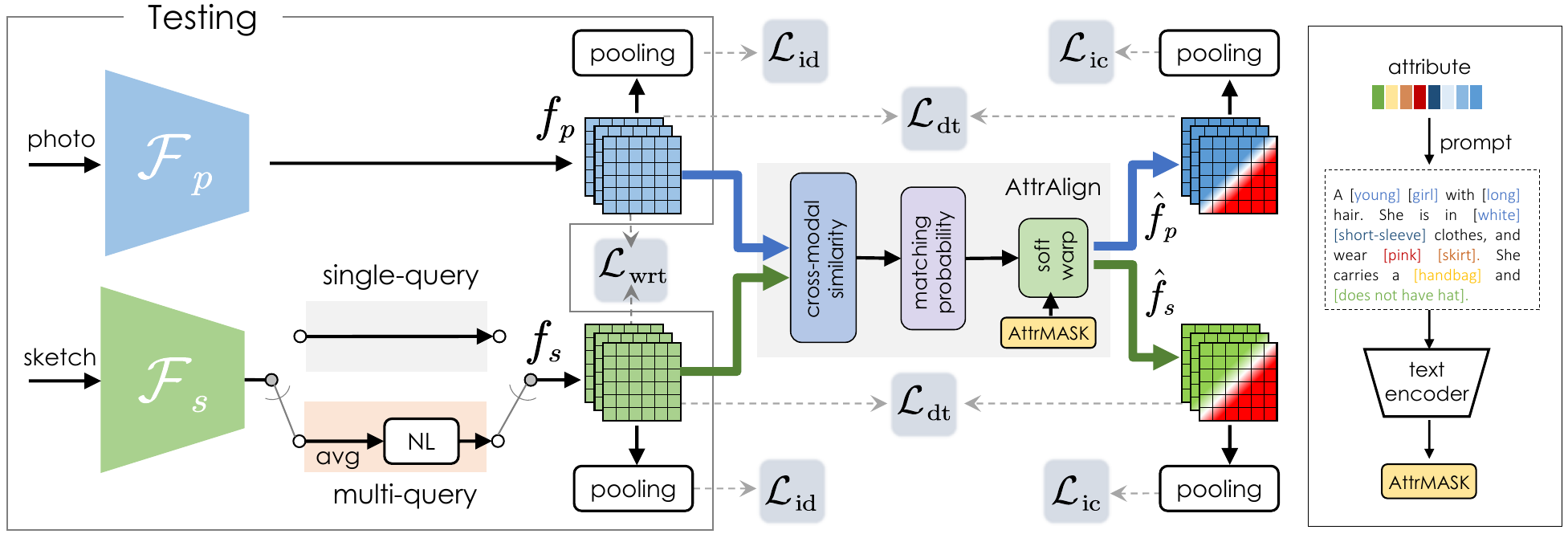}
    \caption{\textbf{Overview of our method.} During training and testing, we can use either single-query or multi-query separately. The framework consists of two feature extractors (see \S\ref{sec:feature extraction}). In multi-query process, a non-local fusion module (NL) is used for fusing features of multiple sketches  (see \S\ref{sec:multi-query fusion}). To align the photo and sketch features, an AttrAlign module is designed (see \S\ref{sec:attribute-guided alignment}), in which AttrMASK is proposed to cast attributes as an implicit mask.
    }
    \label{fig:architecture}
\end{figure*}

Given the query of a target person with either \emph{single} or \emph{multiple} sketches from different witnesses, our goal is to retrieve the corresponding photos of the person in a gallery database. 


For typical cross-modal re-ID, the main focus is to alleviate the modality gap, yet it lacks research on multiple subjective queries with varying query numbers. We plan to tackle this problem threefold: \textbf{a}) scalable sketch input, which adapts to both single-query and multi-query scenarios and is flexible to varying query numbers; \textbf{b}) non-local fusion, for filtering out the noise in multiple queries while capturing long-distance dependencies; \textbf{c}) attribute-guided alignment, utilizing attribute information to counterbalance the subjectivity in sketches.

Our designed architecture is structured as \figref{architecture}.

\subsection{Feature extractor}
\label{sec:feature extraction}
We adopt the dual-path network~\cite{ye2018visible} to separately extract compact feature maps from paired photo $\{x_p^k\}$ and 
\emph{single} sketch $\{{x_s^k}\}$, where $k$ denotes person with identity $y^k$. 
First, we use separate shallow layers to extract domain-specific features.  Then, we adopt a shared fully connected layer on top of the dual-path extractors. 
We denote the embedding function as $\mathcal{F}_p(\cdot)$ for photos and $\mathcal{F}_s(\cdot)$ for sketches.
Therefore, for paired photo
and \emph{single} sketch, the extracted features are represented by

\begin{equation}
    \mathbf{f}_p = \mathcal{F}_p(x_p)\,, \  \ \, \mathbf{f}_s = \mathcal{F}_s(x_s)\,.
\end{equation}

\subsection{Multi-query fusion}
\label{sec:multi-query fusion}

Above, we use only one \emph{single} sketch as query for the photo gallery, which is the same as the previous Sketch
re-ID approaches~\cite{pang2018cross}.
Here, we consider a more practical yet challenging scenario: using \emph{multiple} queries for retrieval.
The fact that people draw different sketches even for the same identity due to their distinct cognition and drawing styles challenges us to effectively fuse reasonable information and filter noisy information. We propose a specific module.

Assume we have multiple sketches $\{{x_s^k}^1, {x_s^k}^2,\dots,{x_s^k}^n\}$ for the person with identity $y^k$, where $n = s^k \ge 1$ denotes the number of sketches from different artists. 
We first apply the embedding function $\mathcal{F}_s$ on individual sketches respectively. 
As step 1 in fusion, to adapt to the varying number of queries, we adopt the parameter-free average pooling (avg) to merge them into a unifying feature map:
\begin{equation}
    x_s^\prime = \textbf{\texttt{avg}}([\mathcal{F}_s({x_s^k}^1) | \mathcal{F}_s({x_s^k}^2) | \dots | \mathcal{F}_s({x_s^k}^n]))\,,
\end{equation}
where $[\cdot|\cdot|\cdot]$ denotes the stacking operation. 
Although the average pooling will compress some valuable information, it helps filter the noisy counterparts. Considering the information loss caused by average pooling, furthermore, we use the non-local attention~\cite{ye2021deep} to obtain the weighted sum of the features at all positions, using both identity-wise and spatial-wise information:
\begin{equation}
    \mathbf{f}_s = \text{NL}(x_s^\prime) = W_{x_s^\prime} * \phi(x_s^\prime) + x_s^\prime\,,\,  
\end{equation}
where $\text{NL}(\cdot)$ denotes the non-local fusion. $W_{x_s^\prime}$ is a weight matrix to be learned. $\phi(\cdot)$ denotes non-local operation~\cite{Wang_2018_CVPR}, which is defined as:
\begin{equation}
    \phi(x(\mathbf{m})) = \frac{1}{\mathcal{C}(x)} \sum_{\forall{\mathbf{n}}} \mathcal{G}(x(\mathbf{m}), x(\mathbf{n})) \mathcal{H}(x(\mathbf{n})),
\end{equation}
where $\mathbf{m}$ and $\mathbf{n}$ is the index of an output position. A pairwise function $\mathcal{G}$ computes a scalar between $\mathbf{m}$ and all $\mathbf{n}$. The unary function $\mathcal{H}$ computes a representation of the input at position $\mathbf{n}$. And the response is normalized by a factor $\mathcal{C}(x)$.
In this way, we can explore and fuse useful information from multiple queries as much as possible.


\subsection{Attribute-guided alignment}
\label{sec:attribute-guided alignment}
Inspired by CMAlign~\cite{park2021learning}, 
we utilize the bidirectional alignment to align photo and sketch features. This module enables the elimination of cross-modal discrepancy. 
Besides the modality discrepancy, subjectivity is the key challenge in Sketch re-ID, different from other cross-modal re-ID tasks~\cite{wang2019beyond}.
To counteract the subjectivity, we introduce objective auxiliary information (\ie attributes) as guidance. The attributes are implicitly cast as a mask to guide the alignment between multi-modal representations. 



\paragraph{Bidirectional alignment.} Firstly, we compute the cross-modal feature similarity between the photo feature $\mathbf{f}_p \!\in\! \mathbb{R}^{h \times w \times d}$ and the sketch feature $\mathbf{f}_s \!\in\! \mathbb{R}^{h \times w \times d}$ as:
\begin{equation}
    C(\mathbf{m},\mathbf{n}) = \frac{\mathbf{f}_{p}(\mathbf{m})^\top \mathbf{f}_{s}(\mathbf{n})}{\Vert\mathbf{f}_{p}(\mathbf{m})\Vert_2 \Vert\mathbf{f}_{s}(\mathbf{n})\Vert_2}\,,
\end{equation}
where $\Vert \cdot \Vert$ denotes the $l2$ normalization function. $\mathbf{f}_I(\mathbf{m})$ and $\mathbf{f}_I(\mathbf{n}$) denote photo and sketch feature of size $d$ at position $\mathbf{m},\mathbf{n}$ respectively, where $I \in \{p,s\}$. Then, we compute the photo-to-sketch matching probability as follows:
\begin{equation}
    P(\mathbf{m},\mathbf{n}) = \frac{\exp(\beta C(\mathbf{m,n}))}{\sum_{\mathbf{n}^\prime}\exp(\beta C(\mathbf{m,n}^\prime))}\,.
\end{equation}
For sketch-to-photo alignment, based on the matching probability, we reconstruct the photo feature $\hat{\mathbf{f}}_p \in \mathbb{R}^{h \times w \times d}$ 
by warping the sketch feature $\mathbf{f}_s$ with a soft warping $\mathcal{W}_p(\cdot)$ as:
\begin{equation}
    \hat{\mathbf{f}}_{p}(\mathbf{m}) = M(\mathbf{m})\mathcal{W}(\mathbf{f}_{s}(\mathbf{m})) + (1-M(\mathbf{m}))\mathbf{f}_{p}(\mathbf{m})\,,
\end{equation}
where $M \in \mathbb{R}^{h \times w}$ denotes the person segmentation map shared by bi-directional alignment. The soft warping function aggregates features using matching probability as 
\begin{equation}\label{eq:softWarp}
    \mathcal{W}(\mathbf{f}_s(\mathbf{m})) = \sum_\mathbf{n} P(\mathbf{m,n})\mathbf{f}_s(\mathbf{n})\,.
\end{equation}
The photo-to-sketch alignment can be derived similarly.

\paragraph{Attribute mask.}
Taking attribute labels are semantically correlated to the corresponding body parts, there exist approaches employing them as labels for direct supervision~\cite{su2016deep, lin2019improving,matsukawa2016person}. Instead, we cast them as a \emph{implicit} mask to align the sketch and photo. Employing the implicit mask instead of the real human mask is because the attribute is accessible from the witness. 
Besides, we introduce natural language supervision~\cite{radford2021learning} motivated by the rapid development of multi-modal learning. 
Concretely, 
\textbf{1)} the attribute labels $\mathbf{a}^k \in \mathbb{Z}^{27}$ are first mapped in to words, and then prompted into a coherent sentence $t^k$ using the chosen template~\cite{lin2019improving}. Here $\mathbf{a}^k$ and $t^k$ denotes the attribute label and text for the corresponding person with identity label $y^k$. For instance, a vector $[2,\dots,1,\dots,2]$ would first be mapped into $["\text{girl}",\dots,"\text{pink}",\dots,"\text{shirt}"]$
, and be inserted into the template. The final text would be “A young \textit{girl} with long hair. She is in white short-sleeve clothes, and wear \textit{pink} \textit{skirt}. She carries a handbag and does not have hat.”
\textbf{2)} We tokenize and project the texts into an image-text shared embedding space, using a pre-trained text encoder from CLIP~\cite{radford2021learning}. The implicit guidance map is computed as follows:
\begin{equation}\label{eq:mask}
    M(\mathbf{m}) = g(\mathbf{f}_t)\,,\,
    \mathbf{f}_t = \mathcal{F}_t(\text{tokenize}(t))\,,
\end{equation}
where $\mathbf{f}_t \in \mathbb{R}^{d}$ denotes the text feature map, $\mathcal{F}_t(\cdot)$ and $g(\cdot)$ denotes embedding function for texts and the projection to align dimensions, respectively.

\subsection{Supervision}

\paragraph{Identity (consistency) loss} It treats the training as a classification problem.
Given an image 
$x$ with label $y$, its probability to be predicted as label $y$ within the total $n$ classes is $p$. 
The probability is calculated from the photo/sketch feature with a pooling layer followed by a softmax function. We then apply cross-entropy function to compute the loss.
The identity loss ($\mathcal{L}_\text{id}$) and identity consistency loss ($\mathcal{L}_\text{ic}$) are computed from the {original} feature and {reconstructed} feature, respectively.

\paragraph{Weighted regularization triplet loss ($\mathcal{L}_\text{wrt}$)} It avoids any additional margin parameter while utilizing the relative distance optimization between positive and negative pairs.

\paragraph{Dense triplet loss ($\mathcal{L}_\text{dt}$)} It locally compares original features from both modalities in pixel-level with a co-attention map (different from standard triplet loss).


\paragraph{Total loss}
Overall, we combine the above objectives together as follows:
\begin{equation}
    \mathcal{L} = \mathcal{L}_\text{id} +
    \mathcal{L}_\text{wrt} + \lambda_\text{ic}\mathcal{L}_\text{ic} +  \lambda_\text{dt}\mathcal{L}_\text{dt}\,,
\end{equation}
where the parameter $\lambda_\text{ic}$ and $\lambda_\text{dt}$ balance dense triplet loss and the ID consistency with others. We set them both to $0.5$. For more detailed explanations of each loss function, please refer to the appendix.



\section{Experiment}

\begin{table}[t]
    \caption{\textbf{Single-query evaluation.} Both train and test are under the single-query setting. $^\dagger$attr means using the attribute feature to query.
    $^\ddagger$$S_k$ denotes randomly sampling a style for each identity, similar to the composition of PKU-Sketch.
    $^\#$$S_\text{all}$ denotes using all sketches in our dataset. 
    We categorize the comparisons into 3 groups of $\{S_1, S_1\}$, $\{S_{all}, S_{all}\}$ and $\{S_k, S_{all}\}$, denoted as G-$S_1$, G-$S_{all}$ and G-$S_k$ respectively according to $\{\text{train}, \text{query}\}$.
    }
    \label{tab:single query comparison}
        \centering
        \tablestyle{4pt}{1.1}
        \resizebox{\columnwidth}{!}{
        \begin{tabular}{l|c|c|ccccc}
              method & train & query &
              \multicolumn{1}{c}{\fontsize{8.5pt}{1em}\selectfont mAP} &
              \multicolumn{1}{c}{\fontsize{8.5pt}{1em}\selectfont r@1} &
              \multicolumn{1}{c}{\fontsize{8.5pt}{1em}\selectfont r@5} &
              \multicolumn{1}{c}{\fontsize{8.5pt}{1em}\selectfont r@10} \\
        \shline
        baseline & $^\dagger$attr & attr & 20.37 & 14.26 & 39.56 & 52.21 \\
        \hline
        {baseline} 
        &  \multirow{3}*{$^\ddagger$$S_{k}$} &  \multirow{3}*{$^\#{S_\text{all}}$} & 11.37 & ~~8.48 & 24.22 & 34.14 \\
        {CMAlign} 
        &&& 11.87 & ~~9.16 & 24.60 & \textbf{35.86} \\
        {Ours}
        &&& \textbf{11.91} & ~~\textbf{9.79} & \textbf{24.89} & 35.06 \\
        \hline
        {baseline} & \multirow{3}*{$S_\text{all}$} & \multirow{3}*{$S_\text{all}$} & 17.48 & 15.02 & 36.03 & 47.43  \\
        {CMAlign}  & & & 18.67 & 16.37 & 37.22 & 49.32   \\
        {Ours} & & & \textbf{19.61} & \textbf{18.10} & \textbf{38.95} & \textbf{50.75} \\
        \hline
        baseline & \multirow{3}*{$S_1$} & \multirow{3}*{$S_1$} & 18.84 & 15.06 & 40.16 & 51.20\\
        CMAlign & & & 19.28 & 20.93 & 39.16 & 54.02 \\
        Ours & & & \textbf{22.89} & \textbf{22.63} & \textbf{40.56} & \textbf{56.63} \\
        \end{tabular}
        }
\end{table}

\begin{table}[t]
    \caption{Comparison with state-of-the-art VI-ReID methods. Both train and test set uses all sketches in our dataset, under single-query setting.}
    \label{tab:single-query VI-ReID}
    \centering
    \tablestyle{4pt}{1.1}
        \resizebox{\columnwidth}{!}{
    \begin{tabular}{r|cccc}
        method & mAP & rank@1 & rank@5 & rank@10\\
        \shline
        DDAG~\cite{ye2020dynamic}  & 12.13 & 11.22 & 25.40 & 35.02\\
        CM-NAS~\cite{fu2021cm}  & 0.82 & 0.70 & 2.00 & 3.90\\
        CAJ~\cite{ye2021channel}  & 2.38 & 1.48 & 3.97 & 7.34\\
        MMN~\cite{zhang2021towards}  & 10.41 & 9.32 & 21.98 & 29.58\\
        DART~\cite{yang2022learning}  & 7.77 & 6.58 & 16.75 & 23.42\\
        DCLNet~\cite{sun2022not}  & 13.45 & 12.24 & 29.20 & 39.58\\
        DSCNet~\cite{zhang2022dual}  & 14.73 & 13.84 & 30.55 & 40.34\\
        DEEN~\cite{zhang2023diverse}  & 12.62 & 12.11 & 25.44 & 30.94\\
        \hline
        Ours (single-query) & \textbf{19.61} & \textbf{18.10} & \textbf{38.95} & \textbf{50.75}\\
        Ours (multi-query) &
        \textbf{24.45} & \textbf{24.70} & \textbf{50.40} & \textbf{63.45}\\

    \end{tabular}
    }
    
\end{table}

\begin{table}[t]
    \caption{\textbf{Multi-query evaluation.} 
    }
    \label{tab:multi-query comparison}
        \centering
        \tablestyle{4pt}{1.1}
        \resizebox{\columnwidth}{!}{
        \begin{tabular}{l|c|cccc}
              method & 
              setting &
              \multicolumn{1}{c}{\fontsize{8.5pt}{1em}\selectfont  mAP} &
              \multicolumn{1}{c}{\fontsize{8.5pt}{1em}\selectfont  r@1} &
              \multicolumn{1}{c}{\fontsize{8.5pt}{1em}\selectfont r@5} &
              \multicolumn{1}{c}{\fontsize{8.5pt}{1em}\selectfont  r@10}\\ 
        \shline
        baseline & \multirow{3}*{ensemble} & ~~0.87 & ~~0.20 & ~~1.00 & ~~2.21 \\
        CMAlign &  & ~~4.15 & ~~1.94 & ~~8.48 & 14.14 \\
        Ours &  & ~~5.39 & ~~3.33 &11.39 & 17.26 \\
        \hline
        baseline & \multirow{3}*{{na\"ive} fusion} & 22.14 & 18.27 & 43.57 & 57.23 \\
        CMAlign &  & 23.19 & 20.48 & 44.98 & 60.24 \\
        Ours & & 23.89 & 22.29 & 49.80 & 59.64 \\
        \hline

         
        Ours & \emph{non-local} &
        \textbf{24.45} & \textbf{24.70} & \textbf{50.40} & \textbf{63.45}\\
        \end{tabular}
        }
\end{table}

\begin{table}[t]
    \caption{\textbf{Testing on unseen styles.} We report the mAP score.}
    \label{tab:unseen}
    \centering
    \begin{minipage}[t]{0.85\linewidth}
        \centering
        \resizebox{1.\linewidth}{!}{
        \begin{tabular}{l|ccccc}
            \multicolumn{6}{c}{\textbf{a}) single-query train and single-query test}\\
            \diagbox{train}{test}& \multicolumn{1}{c}{$S_6$} & $S_{5,6}$ & $S_{4\dots6}$ & $S_{3\dots6}$ & $S_{2\dots6}$\\
            \shline
            $S_1$ & ~~6.72 & ~~4.74 & ~~5.25 & 4.57 & 4.76 \\
            $S_{1,2}$ & ~~8.50 & ~~6.30 & ~~6.14 & 7.78 & -- \\
            $S_{1\dots3}$ & 11.06 & ~~9.42 & 10.30 & -- & -- \\
            $S_{1\dots4}$ & 14.37 & 10.93 & -- & -- & -- \\
            $S_{1\dots5}$ & 14.93 & -- & -- & -- & -- \\
        \end{tabular}}
    \end{minipage}
    \begin{minipage}[t]{0.85\linewidth}
        \centering
        \resizebox{1.\linewidth}{!}{
        \begin{tabular}{l|ccccc}
            \multicolumn{6}{c}{\textbf{b}) multi-query train and multi-query test}\\
            \diagbox{train}{test}
            & $S_{6}$ &
            $S_{5,6}$ & $S_{4\dots6}$ & $S_{3\dots6}$ & 
            $S_{2\dots6}$\\
            \shline
            $S_{1,2}$ & ~~~~--~~~ & ~~7.86 & ~~6.88 & ~~7.75 & ~~~~--~~~ \\
            $S_{1\dots3}$ & ~~~~--~~~ & 10.68 & 10.13 & ~~~~--~~~ & ~~~~--~~~ \\
            $S_{1\dots4}$ & ~~~~--~~~ & 11.85 & ~~~~--~~~ & ~~~~--~~~ & ~~~~--~~~ \\
        \end{tabular}}
    \end{minipage}
    \begin{minipage}[t]{0.85\linewidth}
        \centering
        \resizebox{1.\linewidth}{!}{
        \begin{tabular}{l|cccccc}
            \multicolumn{6}{c}{\textbf{c}) single-query train and multi-query test}\\
            \diagbox{train}{test}&
            $S_6$ &
            $S_{5,6}$ & $S_{4\dots6}$ & $S_{3\dots6}$ & $S_{2\dots6}$ \\
            \shline
            $S_1$ & ~~~~--~~~ & ~~4.93 & ~~5.17 & ~~4.21 & ~~3.93\\
            $S_{1,2}$ & ~~~~--~~~ & ~~5.62 & ~~5.63 & ~~4.27 & ~~~--~~~\\
            $S_{1\dots3}$ & ~~~~--~~~ & ~~6.84 & ~~6.09 & ~~~~--~~~ & ~~~--~~~\\
            $S_{1\dots4}$  & ~~~~--~~~ & ~~8.91 & ~~~~--~~~ & ~~~~--~~~ & ~~~~--~~~ \\
        \end{tabular}
        }
    \end{minipage}
    \begin{minipage}[t]{0.85\linewidth}
        \centering
        \resizebox{1.\linewidth}{!}{
        \begin{tabular}{l|ccccc}
            \multicolumn{6}{c}{\textbf{d}) multi-query train and single-query test}\\
            \diagbox{train}{test}& $S_6$ & $S_{5,6}$ & $S_{4\dots6}$ & $S_{3\dots6}$ & 
            $S_{2\dots6}$\\
            \shline
            $S_{1,2}$ & ~~7.85 & ~~6.02 & ~~5.19 & ~~6.16 & ~~~--~~~\\
            $S_{1\dots3}$ & ~~8.32 & ~~7.09 & ~~7.04  & ~~~~--~~~ & ~~~--~~~\\
            $S_{1\dots4}$ & ~~6.16 & ~~6.23 & ~~~~--~~~ & ~~~~--~~~ & ~~~--~~~\\
            $S_{1\dots5}$ & ~~7.92 & ~~~~--~~~  & ~~~~--~~~ & ~~~~--~~~ & ~~~~--~~~ \\
        \end{tabular}}
    \end{minipage}
\end{table}


\begin{table}[!t]
    \caption{\textbf{Testing on another dataset}. 
    \label{tab:generation}
    $^\dagger$FT denotes fine-tuning.}
    \centering
    \tablestyle{6pt}{1.1}
    \resizebox{1.\linewidth}{!}
    {
    \begin{tabular}{cc|c|cccc}
        train & query & $^\dagger$FT & mAP & r@1 & r@5 & r@10\\\shline
        $S_k$ & pku & - & 45.96 & 46.00 & 74.00 & 82.00\\
        $S_k$ & pku & \checkmark & 67.85 & 70.00 & 88.00 & 94.00\\
        $S_1$ & pku & - & 50.11 & 48.00 & 74.00 & 84.00\\
        $S_1$ & pku & \checkmark & 71.53 & 68.00 & 92.00 & 98.00\\\hline
        $S_\text{all}$ & pku & - & 55.61 & 58.00 & 82.00 & 92.00\\
        $S_\text{all}$ & pku & \checkmark & 78.77 & 78.00 & 96.00 & 98.00\\

    \end{tabular}}
    
\end{table}



\heading{Implementation details.}
Following previous cross-modal re-ID methods~\cite{ye2018visible,lu2020cross,choi2020hi}, we adopt the first residual block of ImageNet-pre-trained~\cite{deng2009imagenet} Resnet50~\cite{he2016deep} as the separate extractor for both modalities, while the rest of Resnet50's parameters are shared. We augment our data with random crops and horizontal flips for training. The dimension of the image and text features are set to 2,048 and 512, respectively. The batch size is 32, with 4 images each for 8 identities. We use Adam optimizer with the same setting in CLIP~\cite{radford2021learning} for optimizing the text encoder. We adopt a warm-up strategy~\cite{luo2019bag} for the SGD optimizer to train the rest of the network. We implement our model with \emph{PyTorch} and train it end-to-end. The random seeds are all set to 0 for a more reliable comparison. For our dataset $S_1$ and $S_\text{all}$, the training takes about 26 and 70 minutes, respectively, with an NVIDIA Tesla V100 16GB GPU.

\subsection{Benchmark results}




\begin{table*}[!tbh]
    \centering
    \caption{\textbf{Ablation study.} Training and testing are
    under the multi-query setting. $^\dagger$\emph{Align} uses the align module without the attribute mask and simply uses a softmax function instead. $^\ddagger$\emph{Noise ratio} $n/27$ means out of the total 27 attributes, $n$ of them are turned into noise randomly.}
    \label{tab:ablation}
    \centering
    \begin{minipage}[t]{0.3\textwidth}
    \tablestyle{4pt}{1.1}
    \begin{tabular}{cccc|cc}
    \multicolumn{6}{c}{\textbf{a}) Loss ablation}\\
    $\mathcal{L}_\text{id}$ & $\mathcal{L}_\text{wrt}$ & $\mathcal{L}_\text{ic}$ & $\mathcal{L}_\text{dt}$ & mAP & r@1\\ \shline
    \checkmark & -- & -- & -- & ~~2.53 & ~~2.01 \\ 
    \checkmark & \checkmark & -- & -- &21.93 & 19.08\\
    \checkmark & \checkmark & \checkmark & -- & 22.79 & 20.28\\
    \checkmark & \checkmark & \checkmark & \checkmark & \textbf{24.45} & \textbf{24.70}\\
    \end{tabular}
    \end{minipage}
    \hfill
    \begin{minipage}[t]{0.3\textwidth}
    \tablestyle{4pt}{1.1}
    \begin{tabular}{ccc|cc}
    \multicolumn{5}{c}{\textbf{b}) Model ablation 
    }\\
    $^\dagger$Align & Attr-Mask & Fusion & mAP & r@1\\ \shline
    -- & -- & -- &  22.14 & 18.27\\
    \checkmark & -- & -- &  23.19 & 20.48\\
    \checkmark & \checkmark & -- & 23.89  & 22.29\\
    \checkmark & \checkmark & \checkmark& \textbf{24.45} & \textbf{24.70}\\
    \end{tabular}
    \end{minipage}
    \hspace{1em}
    \hfill
    \begin{minipage}[t]{0.3\textwidth}
    \tablestyle{4pt}{1.1}
    \begin{tabular}{c|cc}
    \multicolumn{3}{c}{\textbf{c}) Attributes' noise level ablation}\\
         $^\ddagger$Noise ratio & mAP & r@1\\\shline
         0/27 & \textbf{24.45} & \textbf{24.70}\\
         3/27 & 21.29 & 20.28\\
         6/27 & 20.89 & 19.28\\
         9/27 & 20.56 & 16.67
    \end{tabular}
    \end{minipage}
\end{table*}

\paragraph{{Single-query retrieval.}}
We compare our method with the baseline~\cite{ye2021deep} and CMAlign~\cite{park2021learning} on our dataset, as \tabref{single query comparison} shows. We categorize the comparisons into 3 groups. \textbf{1}) Our method achieves the best performance for all metrics in all groups, where we train and query using $S_\text{all}$ (denoted as G-$S_\text{all}$). The results demonstrate the advantage of our customized algorithm. \textbf{2)}~Among all groups, the one that trains and queries using $S_1$ (denoted as G-$S_1$) is superior to the others. 
Compared to G-$S_\text{all}$, G-$S_1$ has fewer training and query samples. It indicates not all sketches are created equally. Therefore, we collect plenty of samples when designing the benchmark enabling a more reasonably justify of algorithms than the PKU-Sketch. Besides, the result should be less reliable when retrieving using only one sketch. Thus, we advocate multi-query retrieval.
\textbf{3}) For each identity, we choose one corresponding sketch \emph{$S_k$} from all available styles, similar to PKU-Sketch. It shows that merely increasing the input diversity cannot make the model generalize better. 
\textbf{4)} \emph{Attr} in baseline means using attributes for retrieval instead of sketches, with features extracted from CLIP (similar to \S\ref{sec:multi-query fusion}). We notice that our method in G-$S_1$ outperforms the attribute-based query. 
It shows that our method uses sketch effectively.

Due to the lack of research in sketch Re-ID and its similarity to VI-ReID,
we conduct experiments using a series of state-of-the-art VI-ReID methods, as shown in \tabref{single-query VI-ReID}. 
Our method using single-query outperformed the best by $4.88\%$ in mAP, demonstrating the superiority of our proposed method. With $9.72\%$ advantage using multi-query, our method shows the significance of incorporating multiple subjective queries.

\setlength{\intextsep}{0pt}%
\setlength{\columnsep}{4pt}%
\begin{wrapfigure}{r}{-4cm}
\vspace{-5mm}
\centering
    
    \tablestyle{4pt}{1.1}
    \begin{tabular}{l|ccc}
    method & \fontsize{8.5pt}{1em}\selectfont mAP & \fontsize{8.5pt}{1em}\selectfont r@1  \\
    \shline
    Triplet SN & -- &  ~~9.0  \\
    GN Siamese & -- & 28.9  \\ 
    CDAFL      & -- & 34.0  \\
    Baseline   & 56.16 & 52.0  \\
    \hline
    Ours & \textbf{66.37} & \textbf{70.0}  
    \end{tabular}

\end{wrapfigure}
We compare our method with leading algorithms designed for cross-modality tasks, namely Triplet SN~\cite{yu2016sketch}, GN Siamese~\cite{sangkloy2016sketchy}, CDAFL~\cite{pang2018cross} on PKU-Sketch. 
We follow the training/evaluation protocol in this dataset and outperform other competing methods.



\paragraph{{Multi-query retrieval.}}
\tabref{multi-query comparison} depicts
the evaluation of different fusion strategies. Compared to the single-query setting (\tabref{single query comparison}), applying {na\"ive} fusion on all methods brings significant improvement.
{Na\"ive} fusion combines the sketch features of each query by a simple average pooling operation. 
Our method with non-local fusion takes global information into account and outperforms any methods we have yet. 
Although we have seen the success of {na\"ive} and non-local fusion, the use of ensemble degrades the performance. To ensemble, we trained our model with 6 styles separately, then extracted sketch features with the best model trained from the corresponding style. 
The results show that high-level fusion such as ensemble cannot incorporate sketches with different styles well. While early fusion such as {na\"ive} fusion fuses features better.

\paragraph{Cross-style retrieval.}
\tabref{unseen} evaluates our model's generalization ability to different styles. We observe that multi-query trained models generate better (\tabref{unseen} a vs b). Besides, when multi-query trained model testing under single-query (\tabref{unseen} b vs d) or single-query trained model testing under multi-query (\tabref{unseen} a vs c), the performance degrades. But the multi-query trained model still dominates the performance (\tabref{unseen} c vs d). 

Except for the cross-style evaluation, we also conduct a cross-dataset evaluation 
 (\tabref{generation}). 
We direct test (without fine-tuning) or fine-tune our model trained with our dataset on PKU-Sketch, following its evaluation protocol. \textbf{1)} We find that direct testing performs comparably or even better than other methods trained on PKU-Sketch. But the direct testing results are still worse than our method trained on PKU-Sketch.  
\textbf{2)} Unsurprisingly, training using all sketches has the best generalization ability.
\textbf{3)} We train our method on $S_k$ and $S_1$ with the same quantity of training samples while $S_k$ has more diverse styles like PKU-Sketch. However, the model trained with $S_1$ beats that trained with $S_k$. It illustrates that training using sketches with mixed styles cannot generalize as well as expected.
\textbf{4)} After fine-tuning, our models pre-trained on $S_1$, $S_k$ and $S_\text{all}$ both exceed that trained on PKU-Sketch, showing the powerful generalization ability.
Significantly, the fine-tuned $S_\text{all}$ model outperforms the PKU-Sketch trained model by a large margin (+12.3 mAP), which is consistent with \tabref{unseen} showing the advantage of the multi-query setting and our tailored model. 
Therefore, we advocate using our multi-query pre-trained model as the initiation for sketch person retrieval.

\subsection{Ablation study} 
\tabref{ablation} depicts the ablation study on our loss, model, and attributes' noise level. \tabref{ablation}(a) illustrates removing one loss at a time. Reducing any loss will get a sub-optimal result.
\tabref{ablation}({b}) shows the effectiveness of our proposed modules. The results in \tabref{ablation}({c}) reveal that the quality of attribute labels affects the alignment performance.
We believe that there are two strategies to avoid the negative effect of noisy attributes. \textbf{1)} During training, double-check attributes obtained with an off-the-shelf attribute detector or annotated manually. \textbf{2)} Remove attributes with high uncertainty.


\section{Conclusions}
\label{sec:conc}


In this paper, we raise a new dataset to highlight the subjectivity in perspective and style variation that exists in Sketch re-ID. 
Our dataset provides room for more flexible retrieval settings, enabling both single-query and multi-query retrieval, along with any scale of sketch inputs for each identity. 
With this more realistic setting, we set three new benchmarks to evaluate the robustness and generalization ability of a proposed algorithm. Our designed novel algorithm achieves leading performance in these benchmarks. 

\heading{Limitations.} 
Although we advocate the multi-query Sketch re-ID, we have to acknowledge that compared with photos, using sketches is much more challenging and less effective (\eg modality gaps, information loss from appearance color, style variation, etc). 
Therefore retrieving a person using a sketch is not as effective as using a photo. Besides, sketches are drawn by humans and based on the witness's memory. 
A gap between the identity's natural appearance and an artist's understanding always exists. 
If obtained target's photo, general re-ID technologies are recommended. 


\heading{Privacy concern.} We should clarify that there is no privacy issue raised by Market-Sketch-1K.  Our dataset is built on a public dataset Market-1501 for research purposes, which doesn't cause privacy issue. And we collected sketches drawn by 6 student artists and obtained their permission to use sketches for academic purposes. 


\begin{acks}
    This work was supported by National Natural Science Foundation of China (62171325), National Key R\&D Project (2021YFC3320301), Hubei Key R\&D Project (2022BAA033), CAAI-Huawei MindSpore Open Fund, JST AIP Acceleration Research JPMJCR22U4, JSPS KAKENHI JP22H03620, JP22H05015, and JP22H00529, and the Value Exchange Engineering, a joint research project between Mercari, Inc. and RIISE.
\end{acks}

{\small
\bibliographystyle{ACM-Reference-Format}
\balance
\bibliography{egbib, add_bib}
}

\appendix
\clearpage
\noindent{\huge{\textbf{Appendix}}}
\vspace{5mm}
\counterwithin{figure}{section}
\counterwithin{table}{section}

This supplemental document provides the details of supervision, the details of evaluation metrics, and additional analysis and comparisons on our dataset.

\section{The details of supervision}

\paragraph{Identity (consistency) loss} treats the training as a classification problem. Given an image 
$x$ with label $y$, its probability to be predicted as label $y$ within the total $n$ classes is $p$. 
The probability is calculated from the photo/sketch feature with a pooling layer followed by a softmax function. Similarly, 
we denote $\hat{p}$ as the probability of the reconstructed feature $\hat{\mathbf{f}}$ to be classified as label $y$. The identity loss and identity consistency loss are computed from the {original} feature and {reconstructed} feature, respectively. It is computed by the cross-entropy
\begin{equation}
    \mathcal{L}_\text{id} = -\frac{1}{n}\sum^n \log(p)\,,\,\mathcal{L}_\text{ic} = -\frac{1}{n}\sum^n \log(\hat{p})\,.
\end{equation}

\paragraph{Weighted regularization triplet loss} avoids any additional margin parameter while utilizing the relative distance optimization between positive and negative pairs
\begin{equation}
\begin{split}
    \mathcal{L}_\text{wrt}(i) = \log(1+\exp(\sum_j w_{ij}^p d_{ij}^p - \sum_k w_{ik}^n d_{ik}^n))\,,\\
    w_{ij}^p = \frac{\exp(d_{ij}^p)}{\sum_{d_{ij}^p \in \mathcal{P}_i} \exp(d_{ij}^p)} \,,\, \ \ \,
    \frac{\exp(-d_{ik}^n)}{\sum_{d_{ik}^n \in \mathcal{N}_i} \exp(-d_{ik}^n)}\,,
\end{split}
\end{equation}
where $(i,j,k)$ represents a hard triplet within each training batch. $\mathcal{P}_i$ and $\mathcal{N}_i$ denotes the positive and negative set for anchor $i$, respectively. $d_{ij}^p$/$d_{ik}^n$ denotes the distance between a positive/negative sample pair.

\paragraph{Dense triplet loss} locally compares original features from both modalities in pixel-level with a co-attention map (different from standard triplet loss), denoted by $A_p \in \mathbb{R}^{h \times w}$ for an image as:
\begin{equation}
    A(\mathbf{m}) = M(\mathbf{m})\mathcal{W}(M(\mathbf{m}))\,,
\end{equation}
where $M(\mathbf{m})$ denotes the implicit guidance mask (\ref{eq:mask}) and $\mathcal{W}(\cdot)$ denotes the soft warping function (\ref{eq:softWarp}). We sample a triplet of anchor, positive and negative images denoted as $a$, $p$, and $n$, where the anchor is in a different modality than the other two. The dense triplet loss is calculated as follows:

\begin{equation}
    \mathcal{L}_\text{dt} = \sum_{I \in \{p,s\}} \sum_{\mathbf{m}} A_I(\mathbf{m})[d_I^+(\mathbf{m}) - d_I^-(\mathbf{m}) + \alpha]_+\,,
\end{equation}
where $\alpha$ is a pre-defined margin and $[\cdot]_+$ indicates $\max(0,\cdot)$. $d_I^+(\mathbf{m})$ and $d_I^-(\mathbf{m})$ compute local distances between an anchor feature and reconstructed ones from positive and negative images, respectively, as follows:
\begin{equation}
\begin{split}
    d_I^+(\mathbf{m}) &= \Vert \mathbf{f}_I^a(\mathbf{m}) - \hat{\mathbf{f}}_I^p(\mathbf{m})\Vert_2 \,,\\
    d_I^-(\mathbf{m}) &= \Vert \mathbf{f}_I^a(\mathbf{m}) - \hat{\mathbf{f}}_I^n(\mathbf{m})\Vert_2 \,.
\end{split}
\end{equation}

\paragraph{Total loss}
Overall, we combine the above objectives together as follows:
\begin{equation}
    \mathcal{L} = \mathcal{L}_\text{id} +
    \mathcal{L}_\text{wrt} + \lambda_\text{ic}\mathcal{L}_\text{ic} +  \lambda_\text{dt}\mathcal{L}_\text{dt}\,,
\end{equation}
where the parameter $\lambda_\text{ic}$ and $\lambda_\text{dt}$ balance dense triplet loss and the ID consistency with others. We set them both to $0.5$.


\section{The details of evaluation metrics}

The mAP evaluates the overall performance and is defined as 
\begin{equation*}\label{eq:map}
     AP = \frac{\sum _{k=1}^n P(k) \times gt(k)}{N_{gt}}, mAP = \frac{\sum _{q=1}^Q AP(q)}{Q},
\end{equation*}
where $k$ is the rank in the recall list of size $n$ and $N_{gt}$ is the number of relevant persons. $P(k)$ is the precision at cut-off $k$ and $gt(k)$ indicates whether the $k$-th recall is correct or not. $Q$ is the number of total query sketches. CMC shows the probability that a query identity appears in different-sized candidate lists. The cumulative match characteristics at rank $k$ are calculated as 
\begin{equation*}\label{eq:map}
     CMC@k = \frac{\sum _{q=1}^Q gt(q,k)}{Q},
\end{equation*}
where $gt(q, k)$ equals 1 when the ground-truth of $q$ sketch appears before rank $k$.

\section{More analysis of our dataset}
\begin{figure*}[!tb]
    \setlength\tabcolsep{4pt}
    \centering
    \includegraphics[width=0.9\linewidth]{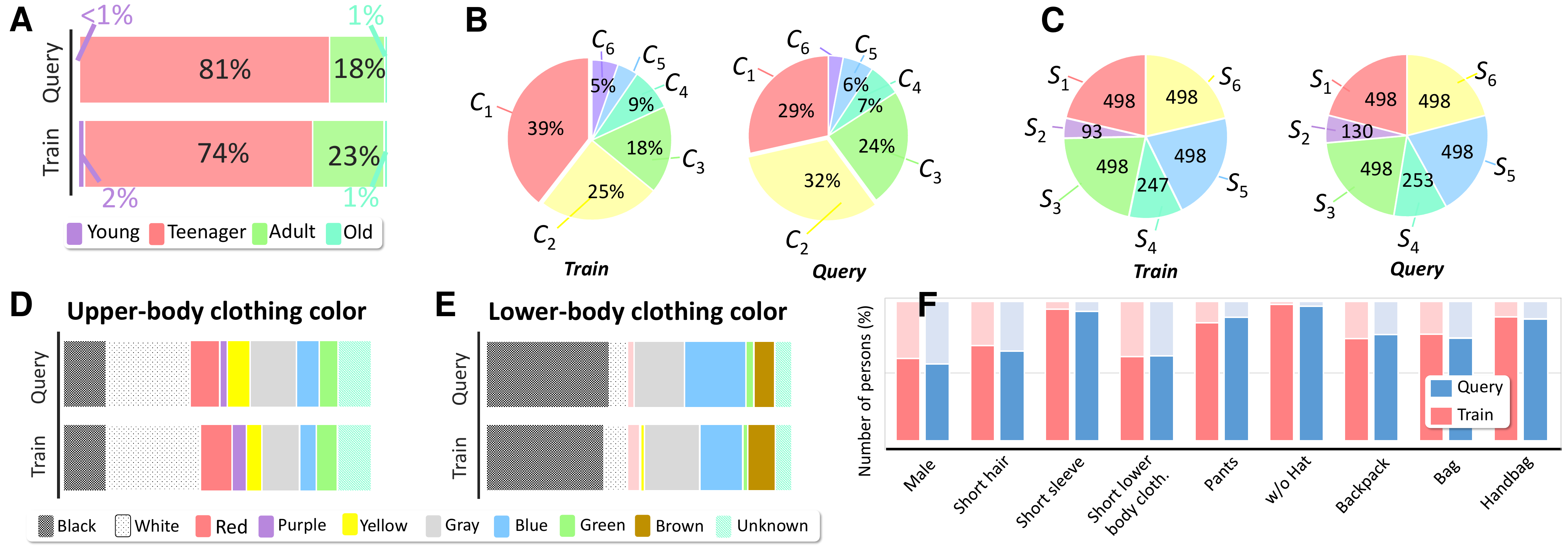}
    \mycaption{Statistics of our dataset}{The figures show the distributions of the training and query subsets selected from the Market-1501 according to the ages~(\textbf{A}), camera views (\textbf{B}), the color of clothes~(\textbf{D}\&\textbf{E}), and other attributes (\eg gender, short/long hair)~(\textbf{F}). \textbf{C}: the distributions of styles of our collected sketches.}
    \label{fig:intro_statPhoto}
\end{figure*}

\figref{intro_statPhoto} presents the statistics of the selected subset. The figures show the diversity of the selected subset. Since we use the subset as the reference, our created sketches are instinctively diverse.

\begin{figure*}[t]
    \centering
    \includegraphics[width=0.9\linewidth]{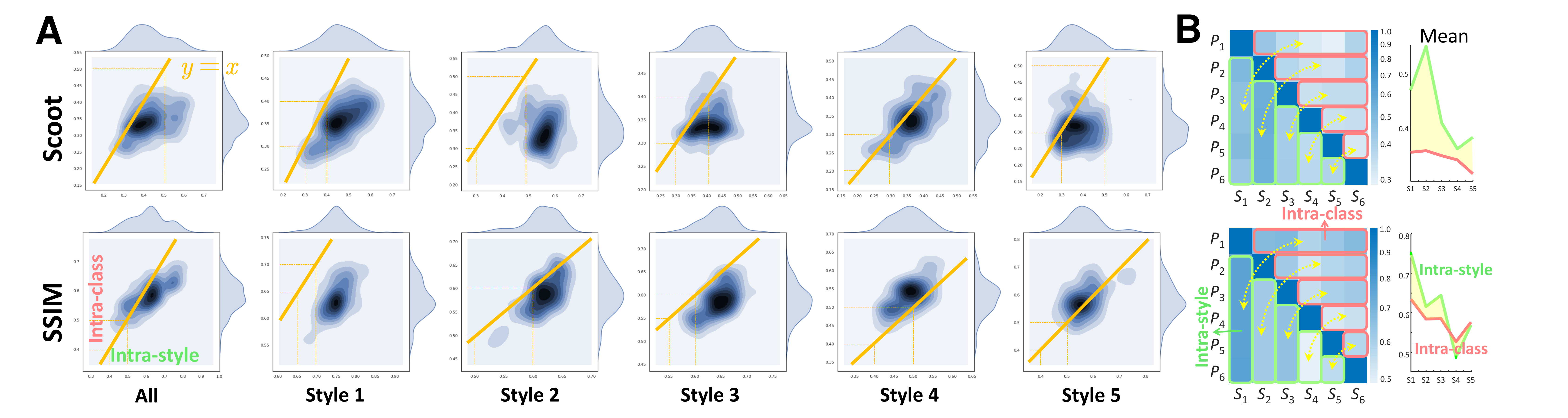}
    \mycaption{Analysis of the sketch similarity}{Two perceptual metrics SSIM and Scoot are used to illustrate the similarity of different sketches. \textbf{A:} the distributions of intra-style and intra-class similarities according to different artists, where the points above/below the line $x\!=\!y$ denote intra-class/intra-style sketches are more similar. \textbf{B:} the similarity matrices of different sketches (The element in row $i$ and column $j$ represents the similarity between the sketch of person $P_i$ by artist $S_i$ and the sketch of person $P_i$ by artist $S_j$.) and the average values of intra-style and intra-class similarities according to different artists. (`intra-style' means sketches are drawn by the same artist, and `intra-class' means sketches are from the same identity.)
    }
    \label{fig:intro_statsketch}
\end{figure*}

\figref{intro_statsketch} shows the similarity between sketches, using two perceptural metrics SSIM and SCOOT. By analyzing the intra-class and intra-style similarity, we can see that in most cases, sketches drawn by the same artist is more similar than those depicting the same target person.

\section{Comparison with related dataset}
We present the statistics of the selected subset. \tabref{compare_more} makes a comparison among different datasets for sketch-based image retrieval tasks. 
A comparison demonstrates 
the superiority of our dataset in the numbers of identities, styles, camera views, photos, sketches, and sketches per style, in particular,
Compared with those for face sketch retrieval and Sketch re-ID, our dataset excels at the number of identities, styles, camera views, etc. as shown in 
\tabref{compare_more}.

\begin{table*}[t]
    \centering
    \caption{Comparisons between related datasets and the created Market-Sketch-1K.}
    \tablestyle{4pt}{1.1}
    \resizebox{\linewidth}{!}{
    \begin{tabular}{l| rrr|rr | rr |rrr | rr |rr}
    {} & \multicolumn{5}{c|}{\textbf{person re-ID}}   & \multicolumn{2}{c|}{\textbf{category}} &\multicolumn{7}{c}{\textbf{instance-level}}\\
    {} & \multicolumn{3}{c|}{{single modality}} & \multicolumn{2}{c|}{{cross modality}} &\multicolumn{2}{c|}{\textbf{-level}} & \multicolumn{3}{c|}{{object sketch}} &\multicolumn{2}{c|}{{face sketch}} & \multicolumn{2}{c}{{person sketch}}\\
    
    {attributes} &VIPeR &Duke  &Market  &SYSU-MM01    &RegDB 
                 &Sketchy &Sketch-250
                 &QMUL-S &QMUL-C &Handbag &VIPSL  & Memory-Aware  &PKU-Sketch  &Ours\\
      \shline
\# ID        &632   &1,404  &1,501 &491      &412    &125     &250     &419    &297    &568    & 200    & 100     & 200     & {996}\\
\# sketch styles per ID       &--    &--     &-- &1        &1      &1       &1       &1      &1      &1      & 5      & 4       & 1       & {6} \\
\# camera views              &2     &8      &6  &6        &2      & --     & --     & --    & --    & --    & --     & --      & 2       & {6} \\
\# color photos              &1264  &36,411 &32,668 &287,628  &4,120  &12,500  &20,000  &2,000  &2,000  &568    & 200    & 10,030  & 400    & {32,668} \\
\# sketches/IR Images & 0    & 0     &0 &15,792   &4,120  &75,471  &204,489 &6,648  &6,730  &568    & 1,000  & 400     & 200     & {4,763} \\
avg. \# sketches per style   &--    &--     &-- &--       &--     &--      & --     &6,648  &6,730  &568    & 200    & 100 &   200       & 793.8 \\
avg. \# sketches per ID      &--    &--     &-- &--       &--     &--      & --     &15.8   &22.6   &1   & 5      & 4       & 1      & 4.78 \\\hline
cross modality               &\xmark&\xmark &\xmark &\cmark   &\cmark &\cmark  &\cmark  &\cmark &\cmark &\cmark &\cmark  &\cmark   &\cmark   &\cmark\\
style factor         &\xmark&\xmark &\xmark&\xmark   &\xmark &\xmark  &\xmark  &\xmark &\xmark &\xmark &\xmark  &\xmark   &\xmark   &\cmark\\
    \end{tabular}
    }
    \label{tab:compare_more}
\end{table*}

\end{document}